# Definition of Visual Speech Element and Research on a Method of Extracting Feature Vector for Korean Lip-Reading

Ha Jong Won, Li Gwang Chol, Kim Hyok Chol, Li Kum Song

College of Computer Science, **Kim Il Sung** University, Pyongyang, DPR of Korea

**Abstract**-In this paper, we defined the viseme (visual speech element) and described about the method of extracting visual feature vector.

We defined the 10 visemes based on vowel by analyzing of Korean utterance and proposed the method of extracting the 20-dimensional visual feature vector, combination of static features and dynamic features.

Lastly, we took an experiment in recognizing words based on 3-viseme HMM and evaluated the efficiency.

**Keywords**- lip-reading, viseme, feature extraction, visual feature vector, HMM

## 1. Introduction

Lip-reading is a process of recognizing the speech from the visual information such as the movement of lip, tongue and tooth.

A lots of researches shows that lip-reading plays an important role in improving the speech recognition of normal people as well as in expressing phonological expression of people with paralyzed auditory organ.

In the middle of 1970s, it was known that lip-reading has limited function that assists auditory with its speech recognition in a noisy environment.

But later researches based of the analysis of the process of speech discrimination of ordinary people has explored that when we see the speaker's face, audio-visual information becomes an part of audio-visual complex improving the encoding of speech signal.

McGruk effect arises when we sense (or hear) the speech signal.[1] It is McGruk effect that is at the bottom of the attempt to use visual information in speech recognition. McGruk effect means that when a man is stimulated by contrary aural and visual stimulation, man recognize nothing. McGruk effect shows that recognized sound depends on not only speech signal but also visual information such as lip movement.

Research on Lip-reading is in progress in 2 parts.

The first one is to assist the improving accuracy of speech recognition and the other one is to recognize the speech contents from only visual information, that is, the movement of lip.

The latter one is more difficult than traditional speech recognition.

In these researches, HMM is used as a recognition model and individual words were set as a basic unit of recognition.

However, Lip-reading systems which don't use speech signal are proper for current number and individual words but, it is difficult to extend word dictionary on a large scale.

Therefore, reference [2] made scientific researches on defining the viseme corresponding phoneme of traditional speech recognition not word as a recognition unit and recognizing words as a sequence of visemes.

A viseme is a virtual speech which has a unique lip shape of mouth so it is similar to phoneme of speech domain. But between phoneme and viseme there exists m to 1 relation, but not the 1 to 1 relation.

Reference [3] described about the definitions of 9 English visemes such as AB,LDF,IDF,LSH,ALF,LLL,RRR,PAL,WWW. LSH Class is classified into small groups: gum-closed nasal speech t, d, n and palate glottis closed - nasal speech g, k, ng.

While reference [4] described about the definitions of 14 English visemes based on the MPEG-4 Multimedia Standard rule, in reference [5] sip and sp, phonemes to represent no sound used in HTK was introduced. And they were classified into 7 consonant visemes, 4 vowel visemes according to the shape of lip formed by vowel and 1 gum sound semivowel viseme to define 13 visemes and to recognize the viseme using HTK.

Viseme can be defined as several ways by using image processing method to get linguistic and visual features. It is "Visemes separated by visual feature" and "linguistic ambiguity of final recognized result" that is important to define viseme.

Lip-reading can convert word sequences into viseme sequences and as ordinary speech recognition based on viseme, i.e., recognition unit, and can get the final sentence using recognition model such as HMM. But when we use lip-reading, it is difficult to get good result as speech recognition because of its own specialty. So reducing the ambiguity came into arise.

Thus, up to now, practical lip-reading system was not developed yet and still have been staying in experiment step was only useful for limited linguistic dictionary

Especially Research on Korean lip-reading is no more than initial step and there are many problems such as the definition and feature extraction of Korean Viseme and designing recognition model.

Visual features for lip-reading can be classified in to two groups, i.e. region feature and contour feature.

A lot of methods such as PCA based region feature extracting method [5], DCT based region feature extracting method [6], region feature extracting method using 2DPCA called BDPCA (Bidirectional PCA)[7], AAM(Active Appearance Model) based lip region feature extracting method [8] and Hi-LDA based region feature extracting method [9] are used to extract region feature.

As region features are got by extracting lip region and applying the transformation such as DCT, PCA or LDA, it is more simple than extracting contour feature, but easy to be affected by illumination and feature vectors extracted using affine transformation such as movement and rotation are much different each other.

There are some methods of extracting lip contour features such as contour feature extracting method applying ASM to extracted lip region [10], after extracting contour of lip, applying affine robust Fourier transform to get Fourier coefficients and to use as feature vector to represent the type of contour [11], a method that gets extracted 10 geometrical parameters from outer and inner contour of lip and use them as a visual feature vector for lip-reading[4] and so on.

The change of lip contour usually reflect the shape features extracted from contour and visual features for lip-reading. So it is more effective than region feature that they represent visual information of speech directly.

But contour based shape features have some disadvantages, that is, the features works properly on assumption that extracted lip contour from lip region should be correct and lip contour can't represent the presence of tooth and the state of tongue itself.

To improve the performance of lip-reading system, visual features reflecting many speech information correctly must be extracted as possible and more efficient recognition algorithm must be used.

In preceding researches, recognition models used in speech recognition were used as the lip-reading used to conduct lip regions extracted from image frames is similar to the process of speech recognition used to conduct speech signals.

Especially, HMM (Hidden Markov Model), the recognition model used in speech recognition whose efficiency was vividly proved are often used.

Whereas in reference [12], it describes about the implementation of lip – reading system based on the HMMs for each viseme, in reference [13], HMMs for each words were formed to use for lip-reading of word.

Reference [14] proposed a method to apply SVM to lip-reading system. SVM is a strong classifier used in pattern recognition such as face detection and face recognition known that it is highly efficient.

In reference [15], it proposed a method of implementing AVSR using TDNN (Time Delayed Neural Network).

In this paper, we put emphasis on the definition of the Korean viseme and extracting visual feature vector.

In Section 2, we define the viseme for Korean lip-reading based on the analysis of Korean Speech.

In Section 3, we describe about the method of extracting static features and dynamic features for Korean lip-reading.

Lastly, Section 4 shows the experimental results for recognizing Korean words using 3 viseme HMM and visual feature vector.

## 2. The definition of Korean Viseme

It is the sub-word that is often used in speech information based speech recognition as a recognition unit.

We can get sub-word by dividing a word into phone and a sub-word is composed of monophone consisted of one phone, biphone depended on the left or right phone, and triphone with left phone and right pone.

In general, sub-word model is are very suitable for composing word model or continuous sentence model, so it is widely used in HMM based speech recognition.

In Korean Speech Recognition, phones are usually defined by Korean consonants and vowels and each word is expressed as the sequence of these phones,

In lip movement based lip-reading system, viseme is defined as a recognition unit and all words are expressed by the sequence of visemes. And 2viseme or 3viseme HMM based recognition model can be configured.

In lip-reading, viseme, a recognition unit, corresponds to a phone in speech recognition and in the case of Korean, it can be defined by Korean consonants and vowels as speech recognition.

But in lip movement based lip-reading, there are many consonants and vowels whose lip movements are similar, so that it is impossible to affirm that there exist 1:1 relation between consonants and vowels and visemes.

Of course, we can define the unique visemes for all Korean consonants and vowels if we consider the states of lip shape, tongue, tooth and vocal cords characterizing utterance of Korean, but in general, it is only lip shape and presence of teeth that can be extracted exactly by image processing.

And it is difficult to extract the others and even though they are extracted, we can't ensure their correctness.

Thus, when we define viseme by only lip shape and presence of teeth, the m: 1 relation, not 1:1 comes into being between consonants-vowels and visemes.

Viseme should be defined to reduce the ambiguity linguistically and to be clear to distinguish as possible.

In general, for the number of visemes, linguistic ambiguity is inversely proportionated to visual division ability. In other words, when we define viseme in detail similarly to consonants and vowels, repetition frequency in viseme representation of words linguistically decreases, but visual recognition accuracy goes down.

In lip-reading for English, we defined 14 visemes discriminated visually and clearly, and matched them with English alphabet.

Unlikely English, each Korean letter is composed of consonant, vowel, and consonant at the end of a syllable. And each consonant, vowel, and consonant at the end of a syllable has certain features visually in lip movement.

A Viseme should contains enough linguistic meaning and it must be possible to discriminate it from another by the image of lip movement.

The utterance of each Korean letter is caused by the activity of different elements composing the human utterance organ (lip, tongue, tooth, vocal cords).

But here, the only information observable visually are shape of mouth, tongue, and state of teeth.

Among these 3 kinds of information, i.e. shape of mouth, tongue and state of teeth, the color information of tongue and lip are so similar that they are indistinguishable.

Thus we must define viseme by using state of teeth and shape of mouse.

Korean letter is composed of consonant + vowel + consonant at the end syllable and here as consonant and consonant at the end of syllable are formed by creating an obstacle to outbreath in mouth and throat, they are mainly determined by the state of tongue rather than shape of mouth, so it is hard to observe.

Among the consonants, it is only "m", "b", "p" and "pp" that are determined clearly by shape of mouth and these consonants have closed shape of mouth.

For example, when we pronounce "ma", after shape of mouth become closed shape, it will be opened shape to pronounce vowel "a".

The 4 consonants above, unlikely the other ones, are the most discriminable ones that can be determined by only shape of mouth.

On the one hand, vowel is mainly discriminated by the difference of shape of mouth. And it is an information visually appeared on the external.

Like this, in general, the shape of mouse and the presence of teeth are changed by Korean vowel and the other consonants and consonants at the end of syllable except "m", "b", "p", "pp" are not expressed by the shape of mouse and the presence of teeth.

Hence, we defined Korean visemes for Korean lip-reading with Korean vowels as its axis as follows. (Table–1)

Table–1. Definition of Korean visemes

| Korean visemes | Denotation |
|---|---|
| a, ya | a |

| o, yo | o |
|---|---|
| u, yu | u |
| i | i |
| e | e |
| we | we |
| wi, ui | wi |
| wa | wa |
| wo | wo |
| m, b, p | m |

Among the visemes above, a, o, i, u, e, m are called "single viseme" and they are determined by one of the basic shapes of mouths as Figure–1.

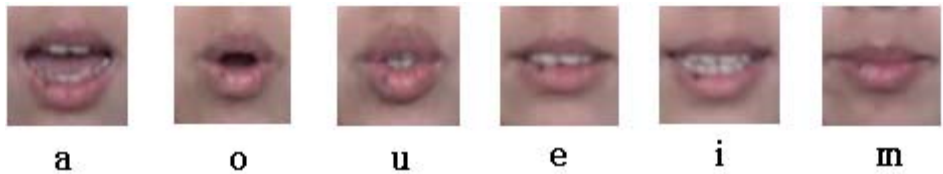

Figure-1. Basic shapes of mouths determining single viseme.

Information about single viseme is mainly reflected to the static feature among static and dynamic features for lip reading, so as a result, static feature about some lip image frame shows which simple viseme that frame represents.

Other hand, the rest 4 visemes we, wi, wa, wo except 6 simple visemes among Korean visemes are called double visemes and they are represented by the combination of 2 simple visemes.

Differently from simple visemes, double visemes are not defined by one main lip shape and are ruled by the dynamic process of 2 main lip shapes.

For example, double viseme 'we' is represented by the combination of simple visemes 'u' and 'e' and is defined by the main lip shape that reflects 'u' and 'e'.

The main lip shapes and moving process which represents double viseme is like as the following Figure-2.

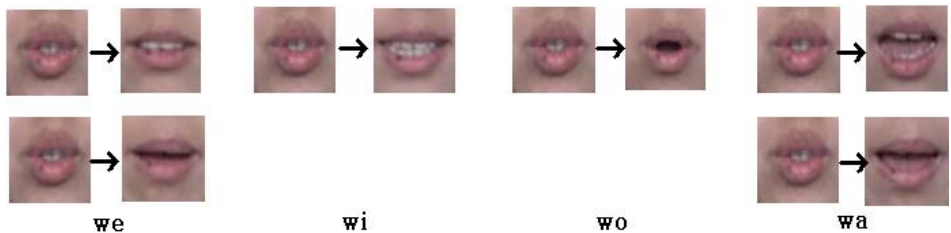

Figure-2. Basic lip shapes that represents double viseme.

The information about double viseme is reflected to the static feature which represents the main lip shape and the dynamic feature which represents the moving processes of the lip shapes.

As above, based on defined Korean viseme, we can express all Korean words as a sequence of visemes.

## 3. Visual feature vector for Korean lip reading

Lip reading, as a process to get the string spoken from the lip moving change, the utterance of each letter is characterized by a certain time delay and moving process at the lip shape.

The moving process of a lip shape when a letter is spoken has 2 characters, that is, one is that main lip shape which specifies the letter exists in the moving process and the other is that this moving process varies from what the four-letter (the letter before) of the letter is.

For example, when a Korean word "Ah" is spoken, we can say that the main lip shape that defines "Ah" is the shape of opened state.

And the main lip shape about "Ah" utters any sentence or doesn't differ according to the speaker but the moving process of lip shape about "Ah" changes according to the front letter of "Ah" in the sentence.

So although the main lip shape of the last state for all Korean letter has almost no change, the moving process of lip shape, dynamic process that characterizes the letter differs from the front letter.

From this, we used the static feature that reflects the main lip shape according to the each letter and dynamic feature that reflects the moving process of lip shape as a visual feature for Korean lip reading.

### 3.1 Static feature

Static feature, a feature related to the lip image frame itself, is a feature that reflect the information that frame got.

We can say that such a information for lip reading is included in lip image region and lip outlet, and is expressed as a distribution of shade values of lip pixels and lip surrounding pixels in lip image region and geometric feature of pixel points that consist the outlet of lip contour.

Therefore, the static feature about all lip image frame can be extracted from the lip image region and the lip outlet.

#### 3.1.1 Region feature

When man is speaking, the lip shape is shown as various shapes according to the speaking letter and the speaker but exists the main lip shape which defines each letter.

In this paper, we defined the 8 main lip shapes (shown in Figure-3) and defined the other lip shapes that is different from the main shapes, as the middle lip-shape that appears during the progress between the main lip shapes and regarded them as those in which the 8 main lip shapes are included in different degrees.

Therefore, every lip shape contains information concerned with all 8 main lip shapes in it and is determined differently according to the degrees of the main lip shapes.

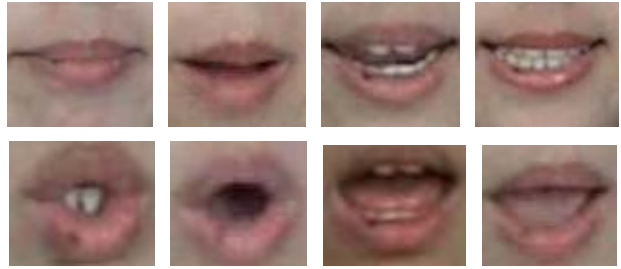

Figure–3. Basic shapes of lip

Thus, regional specification of a lip image is specified by the inclusion degrees how much the 8 main lip shapes are included in it and each degree is defined as the probability- the similarity between the selected lip shape and the main lip shape.

To do this, at first, we have to determine the main probability model $p_i$ ($i = \overline{1, 8}$) about 8 lip shapes.

The main probability mode $p_i$ is determined using the Gaussian Mixing Model (GMM) and is expressed as the followings.

$$p_i(\boldsymbol{x}) = \sum_{k=1}^{K_i} \pi_{ik} g_{ik}(\boldsymbol{x}), \quad i = \overline{1, 8} \tag{1}$$

In previous expression, $K_i$ represents the count of Gaussian component model and $\pi_{ik}$ is the mixing coefficient and is defined as $\sum_{k=1}^{K_i} \pi_{ik} = 1$. And the Gaussian component $g_{ik}$ is expressed as the followings.

$$g_{ik}(\boldsymbol{x}) = \frac{1}{(\sqrt{2\pi})^D \prod_{d=1}^{D} \sigma_{ikd}} \exp\left[-\frac{1}{2}\sum_{d=1}^{D} \frac{(x_d - \mu_{ikd})^2}{\sigma_{ikd}^2}\right] \tag{2}$$

D is the dimension of feature vector x and $\mu_{ikd}, \sigma_{ikd}^2$ are the average and square offset of the Gaussian component $g_{ik}$.

We can get the feature vector for the probability modeling from the lip region image by the following method.

Get $W{\times}H$ DCT coefficients applying 2D-DCT to the lip image in size $W{\times}H$.

In Data compression, low-frequency components, to which the energy is concentrated, is important among the DCT coefficients, but it doesn't mean that those components are also important in Pattern Recognition.

Therefore, in this paper, we extract the coefficients, that contains important information for Pattern Recognition, among all DCT coefficients and use it in probability modeling and select those coefficients in principle of Entropy minimization.

Let's be the set and the size of whole lip images as $S$ and $N$, and the set and size of study lip images concerned to 8 main lip shapes as $S_i$ and $Ni$.

We can get the division entropy $H(S)$ of when other division is done on $S$ by the following expression

$$H(S) = -\sum_{i=1}^{8} \frac{N_i}{N} \ln(\frac{N_i}{N}) \qquad (3)$$

When considering the k-th DCT coefficient, we can divide the train data of $S$ into 8 classes according to the distance minimization principle using this coefficient as a feature valve.

Let $s_i, n_i$ $(i = \overline{1,8})$ be 8 subsets and size got by this division, then for each $S_i$, we can think entropy samely in $S$.

$$H(s_i) = -\sum_{j=1}^{8} \frac{n_{ij}}{n_i} \ln(\frac{n_{ij}}{n_i}), \; i = \overline{1,8} \qquad (4)$$

At the expression above, $n_{ij}$ is the number of study data of j-th main lip class in $S_i$.

Then the division entropy $HD(S, k)$ is like this

$$HD(S,k) = \sum_{i=1}^{8} \frac{n_i}{N} H(s_i) \qquad (5)$$

When we divide $S$ into 8 $S_i$ using $k$th DCT coefficient.

The information gain $IG(k)$ that we can get when we use the $k$th DCT coefficient can be defined as the difference between the entropy $H(S)$ that we calculated before the segmentation of $S$ and the segmentation entropy $HD(S,k)$ that we get after we segmented S into 8 $S_i$ by kth DCT coefficient.

$$IG(k) = H(S) - HD(S,k) \qquad (6)$$

$IG(k)$ is the scale ,that expresses how well we have segmented the train data by kth DCT coefficient, and the bigger that value is the more important related DCT coefficient plays the role in Pattern Recognition.

Then, after determine the information gain $IG(k)$ about $W \times H$ DCT coefficients and rearrange DCT coefficients according to the size of $IG(k)$ and select the biggest d values.

Then, determine the mixing coefficients, the average and the square offset of the main probability model $p_i$ about 8 lip shapes after applying EM algorithm that use $S_i$ as train data set.

The main probability model $p_i$ is the probability model that we dependently found after using $S_i$ as the study set about every type, so it hasn't considered its relativity between the lip shapes.

Therefore, we can use the main probability model $p_i$ and can estimate the final probability model $pr_i$, that considered the relativity between the lip shapes.

$$Pr_i(\mathbf{x}) = \sum_{k=1}^{8} \alpha_{ik} p_k(\mathbf{x}), \quad i = \overline{1,8} \qquad (7)$$

$$\alpha_{ik} = \frac{1}{N_i} \sum_{\mathbf{x} \in S_i} \left[ \frac{p_k(\mathbf{x})}{\left( \sum_{j=1}^{8} p_j(\mathbf{x}) \right)} \right] \qquad (8)$$

The Region feature vector $\mathbf{f}'_{rgn}(n)$ about the lip image of the nth frame can be configured applying d-dimensional vector x, that we can get after applying 2D-DCT to the lip image and multiplying IG value, to expression (5).

$$\mathbf{f}'_{rgn}(n) = \left( \frac{Pr_1(\mathbf{x})}{\overline{Pr_1}}, \frac{Pr_2(\mathbf{x})}{\overline{Pr_2}}, \frac{Pr_3(\mathbf{x})}{\overline{Pr_3}}, \frac{Pr_4(\mathbf{x})}{\overline{Pr_4}}, \frac{Pr_5(\mathbf{x})}{\overline{Pr_5}}, \frac{Pr_6(\mathbf{x})}{\overline{Pr_6}}, \frac{Pr_7(\mathbf{x})}{\overline{Pr_7}}, \frac{Pr_8(\mathbf{x})}{\overline{Pr_8}} \right) \qquad (9)$$

In the above expression, $\overline{Pr_i}$ is the average probability about the i-th lip shape and is defined by $\overline{Pr_i} = \frac{1}{N_i} \sum_{\mathbf{x} \in S_i} Pr_i(\mathbf{x})$.

### 3.1.2 Contour feature

We can get contour feature from the lip contour extracted in every lip image frame.

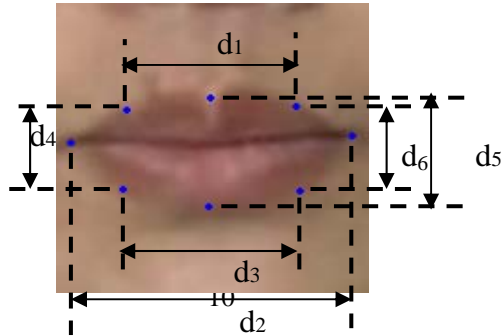

Figure–4. The landmark points used in extraction of the contour feature.

Let landmark points of lip contour taken by n-th lip image frame $pt_i = (x_i, y_i), i = \overline{1,20}$,

In contour feature extraction, only 8 landmark points $pt_1, pt_3, pt_6, pt_9, pt_{11}, pt_{13}, pt_{16}, pt_{19}$ is used and we can get following parameters from $x,y$ positions of this landmark points as shown by Figure-4.

$$d_1 = |x_1 - x_{11}|, d_1 = |x_3 - x_9|, d_3 = |x_{13} - x_{19}|$$
$$d_5 = |y_6 - y_{16}|, d_4 = |y_3 - y_{19}|, d_6 = |y_9 - y_{13}| \quad (10)$$

Contour feature from this parameters are defined by rates of corresponding amount in closed lip.

That is, in closed lip let the values corresponding above $d_1, d_2, d_3, d_4, d_5, d_6$ $d_{1c}, d_{2c}, d_{3c}, d_{4c}, d_{5c}, d_{6c}$ then contour feature for n-th lip image frame $f'_{cont}(n)$ is

$$f'_{cont}(n) = \left( \frac{d_1}{d_{1c}}, \frac{d_2}{d_{2c}}, \frac{d_3}{d_{3c}}, \frac{d_4}{d_{4c}}, \frac{d_5}{d_{5c}}, \frac{d_6}{d_{6c}} \right) \quad (11)$$

Contour feature $f'_{cont}(n)$ takes the relative values to closed lip and this has nothing to do with frame size and has features that reflect the variation of lip contour.

Here, setting the closed lip to the basis is because lip keeps closed usually when the human doesn't speak or when there is no-sound.

To calculate contour feature $f'_{cont}(n)$, we must get 6 parameters $d_{kc} (k = \overline{1,6})$ for closed lip first.

To do this, we must get closed lip from any video but maybe the frame that has closed lip shape appear in the video or not. So consider 2 cases to get parameters for closed lip.

First, in case that closed lip appear in the video, we can extract the closed lip from the video and estimate the parameters directly.

We can determine from the values of region feature whether the lip in every frame of the video is closed or not.

That is, let the corresponding component to closed lip shape among 8-d region feature vector i-th component. Then if it is greater than i-th component $f_i$ extracted from lip region of a frame then this lip of the frame is decided by closed lip.

Then the parameter for the closed lip $d_{kc}, k = \overline{1,6}$ is setted by average of $d_k^t$ which is getted from the frames of which $f_i$ is greater than 1 respectively.

That is

$$d_{kc} = \frac{1}{\left[\sum_{t=1}^{T} sgn(f_i^t - 1)\right]} \sum_{t=1}^{T} \left[sgn(f_i^t - 1) \times d_k^t\right], k = \overline{1,6} \quad (12)$$

In the above expression $T$ is frame count in the video, $f_i^t$ is i-th component of region feature $f'_{rgn}(t)$ extracted from t-th frame.

And $sgn(x)$ is sign function and $sgn(x) = \begin{cases} 1, & x > 0 \\ 0, & x \leq 0 \end{cases}$

In case that there is no closed lip in the video, i-th component $f_i^t$ of region feature is smaller than 1.

In this case, the parameters for the closed lip is estimated from other not closed lips indirectly.

Then the parameters for closed lip $d_{kc}$ is determined by the following expression (13).

$$d_{kc} = \gamma_k^j \overline{d_k}, k = \overline{1,6} \quad (13)$$

In the above expression $\gamma_k^j$ shows dimension rate between k-th parameter of j-th lip-shape which is not closed lip among 8 lip-shapes and k-th parameter of closed lip and comes from statistical analyisis of training data of lip-shapes.

And $\gamma_k^j$ is unique parameter for a talker that is set the other value as the lip shape by one fixed talker and that is different according to a talker.

Static feature vector $f'_{stc}(n)$ for n-th frame is 14-d vector combined with 8-d region feature vector and 6-d contour feature vector

$$f'_{stc}(n) = (s_{rgn} \times f'_{rgn}(n), s_{cont} \times f'_{cont}(n)) \quad (14)$$

In the above expression $s_{rgn}$, $s_{cont}$ is respectively scale transform factor for the region feature and the contour feature.

Region feature in static feature reflects the distribute state for the feature pixels such as the teeth but contour feature reflects the geometrical feature of the landmark points of which lip contour consists such as lip-open extent.

And To reduce the noise effect, the final static feature vector $f_{stc}(n)$ of n-th lip image frame is averaged static feature vector of all frames in the window region which is $W+1$ wide and centered by n-th frame.

$$f_{stc}(n) = \frac{1}{(W+1)} \sum_{k=-W/2}^{W/2} f'_{stc}(n+k) \qquad (15)$$

## 3.2 Dynamic feature

Static feature reflects lip shape of every frame but dynamic feature reflects variation progress of lip shape between from one frame to another frame.

Specially, Dynamic feature is variation of lip shape between neighbor frames centered by current frame, that is, it reflects front-back context so it is very important on reading sequence using triviseme HMM.

Figure-5 shows how lip shape change by left-right context when the human speaks.

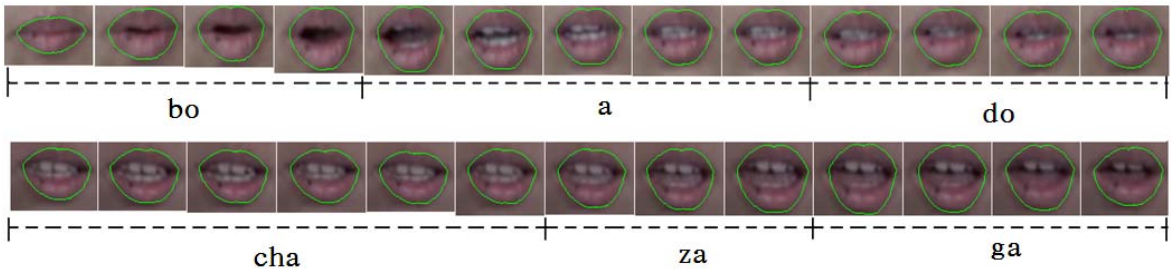

Figure–5. Variation Example of lip shape by left-right context

This shows that the basic lip shape which defines a character is different according to a speaker and a left-right context of the character.

The basic lip shape which defines a character is different according to a speaker and a left-right context but in the array of lip images lip shape the variation feature is not different a lot when the different speakers speak the same sentence (word) but it is different according to which sentence is spoken.

As you can see the above example, as you see the array of lip images for the spoken sentence "Boado", the lip shape, the change progress of lip shape, is "Round lip shape"->"opened lip shape->"Round lip shape" and for "Chazaga", the change progress of lip shape is "bit opened lip shaped"->"large opened lip shape". As a result it is different according to the spoken sentence and it is independent on a speaker.

This shows that the feature reflected lip shape change progress plays very important role to construct the reading sequence system which is independent on a speaker and to perform lip reading using triviseme HMM considering the left-right context.

Dynamic feature reflecting the change progress of lip shape is extracted from transform information of lip contour according to time.

Dynamic feature is defined by position variation according to time of 8 feature points among the 20 landmark points of which lip contour is consisted as the figure-4.

Among the 8 landmark points, for $pt_1$ and $pt_{11}$ corresponding to the corner point of the lip contour the variation of only x-direction is considered, but for $pt_6$ and $pt_{16}$ the variation of only y-direction is considered.

And for the rest 4 landmark points $pt_3$, $pt_9$, $pt_{13}$, $pt_{19}$ the variation of both of $x$, $y$ direction is considered and totally they consist 12-d dynamic feature.

Now let the $x$, $y$ coordinate of landmark point $pt_k$ according to $t$ express $C_k(x, t)$, $C_k(y, t)$ then dynamic feature $f'_{dyn}(t)$ at the time $t$ is expressed by derivation value of correspond $x$, $y$ coordinate of time $t$.

$$f\grave{}_{dyn}(t) = \left(\frac{dC_1(x,t)}{dt}, \frac{dC_3(x,t)}{dt}, \frac{dC_3(y,t)}{dt}, \frac{dC_6(y,t)}{dt}, \frac{dC_9(x,t)}{dt}, \frac{dC_9(y,t)}{dt}\right.$$
$$\left.\frac{dC_{11}(x,t)}{dt}, \frac{dC_{13}(x,t)}{dt}, \frac{dC_{13}(y,t)}{dt}, \frac{dC_{16}(y,t)}{dt}, \frac{dC_{19}(x,t)}{dt}, \frac{dC_{19}(y,t)}{dt}\right)$$
(16)

Input object is the array of lip image so the derivation of $x$, $y$ coordinate at time $t$ is simply the variation of $x$, $y$ coordinate between n th frame and n-1 th frame. That is

$$\frac{dC_k(x,t)}{dt} = \Delta C_k(x,n) = C_k(x,n) - C_k(x,n-1)$$
$$\frac{dC_k(y,t)}{dt} = \Delta C_k(y,n) = C_k(y,n) - C_k(y,n-1)$$
(17)

And dynamic feature $f'_{dyn}(n)$ on the n-th frame is

$$f\grave{}_{dyn}(n) = \left(\Delta C_1(x,n), \Delta C_3(x,n), \Delta C_3(y,n), \Delta C_6(y,n), \Delta C_9(x,n), \Delta C_9(y,n)\right.$$
$$\left.\Delta C_{11}(x,n), \Delta C_{13}(x,n), \Delta C_{13}(y,n), \Delta C_{16}(y,n), \Delta C_{19}(x,n), \Delta C_{19}(y,n)\right)$$
(18)

But actually the variation of lip contour between the neighbor frames in the lip video is very small and the higher the fps of the camera is the smaller this variation gets.

And it is affected by the noisy a little detecting the lip contour so we don't extract the dynamic feature by every frame as a unit but extract the dynamic feature by the frame win dow as a unit, the window is the pack of the frames which is $W+1$ wide.

Then dynamic feature $f_{dyn}(n)$ is defined as the difference between the averaged $x$, $y$ coordinate in $n$th window and the averaged $x$, $y$ coordinate in $n$-1th frame pack.

$$f_{dyn}(n) = \left(\Delta \overline{C_1}(x,n), \Delta \overline{C_3}(x,n), \Delta \overline{C_3}(y,n), \Delta \overline{C_6}(y,n), \Delta \overline{C_9}(x,n), \Delta \overline{C_9}(y,n)\right.$$
$$\left.\Delta \overline{C_{11}}(x,n), \Delta \overline{C_{13}}(x,n), \Delta \overline{C_{31}}(y,n), \Delta \overline{C_{16}}(y,n), \Delta \overline{C_{19}}(x,n), \Delta \overline{C_{19}}(y,n)\right)$$
(19)

Here,

$\Delta \overline{C_k}(x,n) = \overline{C_k}(x,n) - \overline{C_k}(x,n-1)$. And $\overline{C_k}(x,n), \overline{C_k}(y,n)$ is the average of *x*, *y* of *k* th landmark point of every frame in the *n*th window. That is

$$\overline{C_k}(x,n) = \frac{1}{(W+1)} \sum_{k=-W/2}^{W/2} C_k(x, n+k), \quad \overline{C_k}(y,n) = \frac{1}{(W+1)} \sum_{k=-W/2}^{W/2} C_k(y, n+k) \quad (20)$$

## 4. Experiment Result

In this paper, we evaluate the recognize performance of Korean lip reading system in the different environment and analysis of this.

### 4.1 Recognition performance evaluation according to utterance unit

In this paper, we evaluate the propriety of suggested visual feature by recognizing a isolated word which utterance unit of speaker has a word.

First, we defined triviseme HMMs based on single visemes defined in Sec.2 and estimated the models using HTK3.4.

Train data for estimation of triviseme HMMs is the feature vector data extracted from 5, 6000 lip motion video spoken about standard training sentences. Video is the video spoken the standard study sentences by 7 speaker and it is the front faced video.

To evaluate recognize performance of the isolated word we used 300 words which is in the training sentences and 200 words which is not in the training sentences.

In case of N-best recognition, the recognition result is evaluated as truth when 3 candidates result include correct word .

We also took recognition in cases when visemes were given manually, and also when they were not given. The recognition result is as following table.

Table–2. Recognition result according to utterance unit

|  | Recognition accuracy(%) | |
| --- | --- | --- |
|  | Manual viseme information | No manual viseme information |
| Training word | 65.4 | 89.7 |
| Non-training word | 56.5 | 78.9 |

In this table, you can see that recognition accuracy of training words is better than of non-training words.

And you can also know that the more definite visual speech element for word was given, the higher accuracy the result is.

As result of test about scale of word dictionary, the larger the dictionary is, the lower accuracy is.

In the case of recognition of isolated word, language model can't be apply and recognition proceed only by triviseme HMM, the accuracy is low and it's only for validation test of accuracy of 3 visual speech element.

**4.2 Evaluation of Recognition according to speaker**

In the thesis, we have experiments about different speakers for robustness of visual feature vector which we proposed.

For that, we made 50 lip videos of every 3 speakers, one person is training speaker and the others are non-training speakers. And then we have recognition test with 150 lip videos.

At that time, we have given same viseme information for same sentence of every speakers.

The result of recognition experiment for speakers are as follow.

Table–3. Recognition result according to speaker

|  | Recognition accuracy(%) | |
| --- | --- | --- |
|  | Manual viseme information | No manual viseme information |
| Training speaker | 52.5 | 92.3 |
| Non-training speaker1 | 48.1 | 83.8 |
| Non-training Speaker2 | 47.6 | 82.1. |

As you can see in the table, the recognition accuracy for the non-training speaker is approximately 9~10% lower than study speaker.

As the result of analyzing visual feature for the 3 speakers, the variation of the static feature is large but the variation of the dynamic feature is not large for the speaking of the same sentence.

This shows that the basic lip shape is different according to the speaker but the variation feature of the lip for the spoken content is not different a lot.

Therefore we can see the stronger robustness of the dynamic feature than the static feature.

As the result of the above experiment we verified the validity of the Korean visemes which is defined in this paper and the effectiveness of the visual feature vector.

In the future, we are going to concentrate on the selection of reasonable visual features and enhancement of the robustness in the preprocessing step, and the research of effective train algorithm to reduce the train time in training step.

And in the recognize section we are going to intensify the research about effective using the dictionary related spoken content in the recognize progress and the design method of advanced language model.

IDENTIFICATION BY LIPREADING", Dept.of Electronic and Electrical Engineering University of Sheffield,Sheffield S1 3JD,UK, 1997

[12] Hong-xun Yao, Wen Gao1 and Wei Shan," Visual Features Extracting & Selecting for Lipreading", AVBPA 2003, LNCS 2688,pp.251-259, 2003

[13] Kazuhiro Nakamura,Noriaki Murakami,Kazuyoshi Takagi and Naofumi Takagi,"A Real-time Lipreading LSI for Word Recognition", IEEE,2002,pp.301-306

[14] Mihaela Gordan,Constantine Kotropoulos and Ioannis Pitas,"Application of Support Vector Machines Classifiers to Visual Speech Recognition",IEEE ICIP 2002,pp.129-132

[15] D.G.Stork,G.Wolf and E.Levine,"Neural network lipreading system for improved speech recognition",IJCNN,1992.6